\documentclass[letterpaper, 10 pt, conference]{ieeeconf} 
\IEEEoverridecommandlockouts
\makeatletter
\let\NAT@parse\undefined
\makeatother

\usepackage{subcaption}
\usepackage{textcomp}
\usepackage{xcolor}
\usepackage{wrapfig}
\usepackage{dirtytalk}
\usepackage[colorlinks=true,linkcolor=blue,urlcolor=blue,citecolor=blue]{hyperref}
\usepackage{tabularx}
\usepackage[ruled, linesnumbered]{algorithm2e}
\usepackage{booktabs}
\usepackage{multirow}
\usepackage{csquotes}
\usepackage{booktabs} 
\usepackage{makecell}
\usepackage{threeparttable, tablefootnote}
\usepackage{graphicx}
\usepackage{cite}
\usepackage{amsmath,amssymb,amsfonts}
\usepackage{algorithmic}
\usepackage{comment}
\usepackage{balance}
\usepackage{pdflscape,booktabs,tabularx,ragged2e}
\def\BibTeX{{\rm B\kern-.05em{\sc i\kern-.025em b}\kern-.08em
    T\kern-.1667em\lower.7ex\hbox{E}\kern-.125emX}}
\usepackage[framemethod=TikZ]{mdframed}
\let\labelindent\relax
\usepackage{enumitem}
\begin{document}
\title{\LARGE \bf
High Throughput Phenotyping of Physician Notes with Large Language and Hybrid NLP Models}
\author{ Syed I. Munzir, Daniel B. Hier, Michael D. Carrithers
        \thanks{Syed Munzir, Daniel Hier, and Michael Carrithers are with the Department of Neurology and Rehabilitation,  University of Illinois at Chicago, Chicago, USA. Email: \{smunz2, dhier, mcar1\}@uic.edu}
        \thanks{Dr. Carrithers is an employee of the
Department of Veterans Affairs and receives research funding from a VA Merit Award (BLR\&D BX000467).  The content is solely the responsibility of the authors and does not
represent the official views of the US Department of Veterans Affairs. \textbf{This work has been submitted to the IEEE for possible publication.} Copyright may be transferred without notice, after which this version may no longer be accessible}
        }
\maketitle
\thispagestyle{empty} 
\pagestyle{empty} 

\begin{abstract}
Deep phenotyping is the detailed description of patient signs and symptoms using concepts from an ontology.  The deep phenotyping of the numerous physician notes in electronic health records requires high throughput methods. Over the past thirty years, progress toward making high throughput phenotyping feasible. In this study, we demonstrate that a large language model and a hybrid NLP model (combining word vectors with a machine learning classifier) can perform high throughput phenotyping on physician notes with high accuracy. Large language models will likely emerge as the preferred method for high throughput deep phenotyping of physician notes.

{\textit{Clinical relevance:}
Large language models will likely emerge as the dominant method for the high throughput phenotyping of signs and symptoms in physician notes.} 
\end{abstract}
\section{Introduction}
Precision medicine entails the {precise} matching of diagnosis, outcome, treatment, and molecular mechanism of disease to individual patients \cite{sahu2022artificial, afzal2020precision}. An important component of precision medicine is the precise description of the signs and symptoms of disease with standardized and computable terms from a suitable ontology \cite{robinson2012deep}. This detailed description of signs and symptoms has been called \textit{deep phenotyping}. The deep phenotyping of thousands or even millions of physician notes within electronic health records requires high throughput methods \cite{hier2022focused, robinson2012deep, hier2022dc}. Although high throughput phenotyping has gained traction in agriculture \cite{mir2019high, gehan2017high}, its adoption in human medicine has been slower \cite{hier2022dc}. Improved methods for high throughput phenotyping of electronic health records are needed for precision medicine \cite{alzoubi2019review, shivade2014review, pathak2013electronic}.

The extraction of signs and symptoms from electronic health records is a two-step process that involves identification and mapping. \textit{Identification} finds spans of text that describe the signs and symptoms of the disease, which are then \textit{mapped} to concepts in a suitable disease ontology and assigned a machine-readable code\cite{krauthammer2004term}. Concept extraction (sometimes called concept recognition or concept identification) is closely related to the natural language task of named entity recognition (NER) \cite{Bird, luque2019advanced, ahmad2023review}. Medical concepts in free text can be extracted manually, but this work is slow, laborious, and tedious.

\begin{mdframed}[
  frametitle={Box 1: Instructions to GPT-4 and the human annotator for neurological phenotyping},
  frametitlebackgroundcolor=gray!30, 
  backgroundcolor=gray!20, 
  frametitlerule=true,
  frametitlealignment=\centering,
  innertopmargin=10pt,
  innerbottommargin=10pt,
  innerleftmargin=10pt,
  innerrightmargin=10pt,
  roundcorner=10pt,
  font=\footnotesize 
]

Find neurological signs (what a physician finds) and symptoms (what a patient complains of) in physician notes. Ignore radiological, laboratory, EMG, EEG, or pathology findings. Signs and symptoms may be single words such as \underline{ataxia} or short phrases such as \underline{lack of coordination}. Include past and present signs and symptoms. Only report abnormal signs and symptoms and ignore normal and negative findings. After identifying signs and symptoms, each sign and symptom should be assigned to one of these phenotype categories:

\begin{itemize}[ left=0pt, itemsep=1pt]
  \item \underline{behavior}: including anxiety, depression, delusion, psychosis, hallucination, etc.
  \item \underline{cognitive}: memory loss, forgetfulness, confusion, cognitive impairment, inattention, dementia, etc.
  \item \underline{EOM}: including double vision, abnormal eye movements, diplopia, sixth nerve palsy, third nerve palsy, skew deviation, etc.
  \item \underline{fatigue}: including tiredness, lack of energy, poor energy, etc.
  \item \underline{gait}: including abnormal gait, spastic gait, ataxic gait, poor balance, imbalance, falling, falling, falls, uses a cane, uses walker, uses wheelchair, etc.
  \item \underline{hyperreflexia}: including increased reflexes, increased biceps reflex, biceps +++, triceps +++, biceps ++++, triceps +++,etc.
  \item \underline{hypertonia}: increased tone, spasticity, hypertonia, muscle spasms, etc.
  \item \underline{hyporeflexia}: including decreased reflexes, areflexia, hyporeflexia, bicep 1+ 1+, knee 1+ 1+, absent ankle reflex, ankle reflex 1+ 1+, etc.
  \item \underline{sphincter}: including urinary frequency, urinary incontinence, constipation, bowel incontinence, urinary retention, etc.
  \item \underline{incoordination}: including ataxia, dysmetria, poor coordination, incoordination, etc.
  \item \underline{ON}: including optic neuritis, apd, afferent pupillary defect, pale disk, disk atrophy, etc.
  \item \underline{pain}: including shooting pain, burning pain, allodynia, arm pain, headache, head pain, leg pain, etc.
  \item \underline{paresthesias}: including numbness, tingling, loss of sensation, sensory loss, hypesthesia, etc.
  \item \underline{seizure}: including seizures, convulsions, fits, attacks, etc.
  \item \underline{sleep}: including hypersomnia, insomnia, restless legs, abnormal sleep, trouble sleeping, etc.
  \item \underline{speech}: lack of speech, slurred speech, dysarthria, aphasia, etc.
  \item \underline{tremor}: including tremor, tremulousness, action tremor, resting tremor, rubral tremor, etc.
  \item \underline{vision}: including impaired vision, decreased visual acuity, visual loss, etc.
  \item \underline{weakness}: including weakness, loss of strength, difficulty lifting arms, difficulty lifting legs, biceps 4 4, triceps 4 4, hip flexors 4 4, hip extensors 4 4, etc.
\end{itemize}
\end{mdframed}

Natural language processing (NLP) has been searching for methods to automate the extraction of medical concepts from free text. Automation of medical concept extraction makes high throughput phenotyping of electronic health records feasible. Progress towards this goal has been significant, and powerful fifth-generation systems are emerging. 

\begin{enumerate}
    \item \textbf{First-Generation:} Dictionary-based and rule-based systems \cite{krauthammer2004term, eltyeb2014chemical, quimbaya2016named}.
    \item \textbf{Second-Generation:} Machine learning algorithms like conditional random fields, support vector machines, and hidden Markov models \cite{hirschman2002rutabaga, uzuner20112010}. Other second-generation systems made important advances in linguistic analysis \cite{aronson2010overview, savova2010mayo}.
    \item \textbf{Third-Generation:} Deep learning models, particularly those using RNNs (Recurrent Neural Networks) and CNNs (Convolutional Neural Networks) \cite{lample2016neural,chiu2016named,  habibi2017deep, gehrmann2018comparing, arbabi2019identifying}
    \item \textbf{Fourth-Generation:} BERT (Bidirectional Encoder Representations from Transformers) and transformer-based models, introducing the self-attention mechanism. A shift from static word embeddings to context-aware word embeddings that facilitate a deeper understanding of language\cite{devlin2018bert, vaswani2017attention, zhu2021utilizing, yu2019biobert, lee2020biobert, ji2020bert}.
    \item \textbf{Fifth-Generation:} General-purpose large language models (LLMs) like GPT-4 (Generative Pre-trained Transformer 4), which build upon the transformer architecture to provide more versatility, scalability, and generalizability across a wide range of tasks.
\end{enumerate}

General-purpose large language models, such as GPT-4, are emerging that can perform difficult NLP tasks such as the phenotyping of physician notes \cite{yan2023large, yang2023enhancing,wang2023fine} without additional model training.

\textbf{Experimental Setup.} We manually annotated the signs and symptoms (phenotype) in the physician notes of thirty multiple sclerosis patients. Notes were phenotyped by a hybrid model (\textit{NimbleMiner})  and a general-purpose large language model (\textit{GPT-4}). The phenotypes created by the two models were compared to the ground truth labels. Performance metrics, including accuracy, precision, recall, specificity, and precision, were calculated for each model.

\begin{figure}
    \centering
    \includegraphics[width=0.5\textwidth]{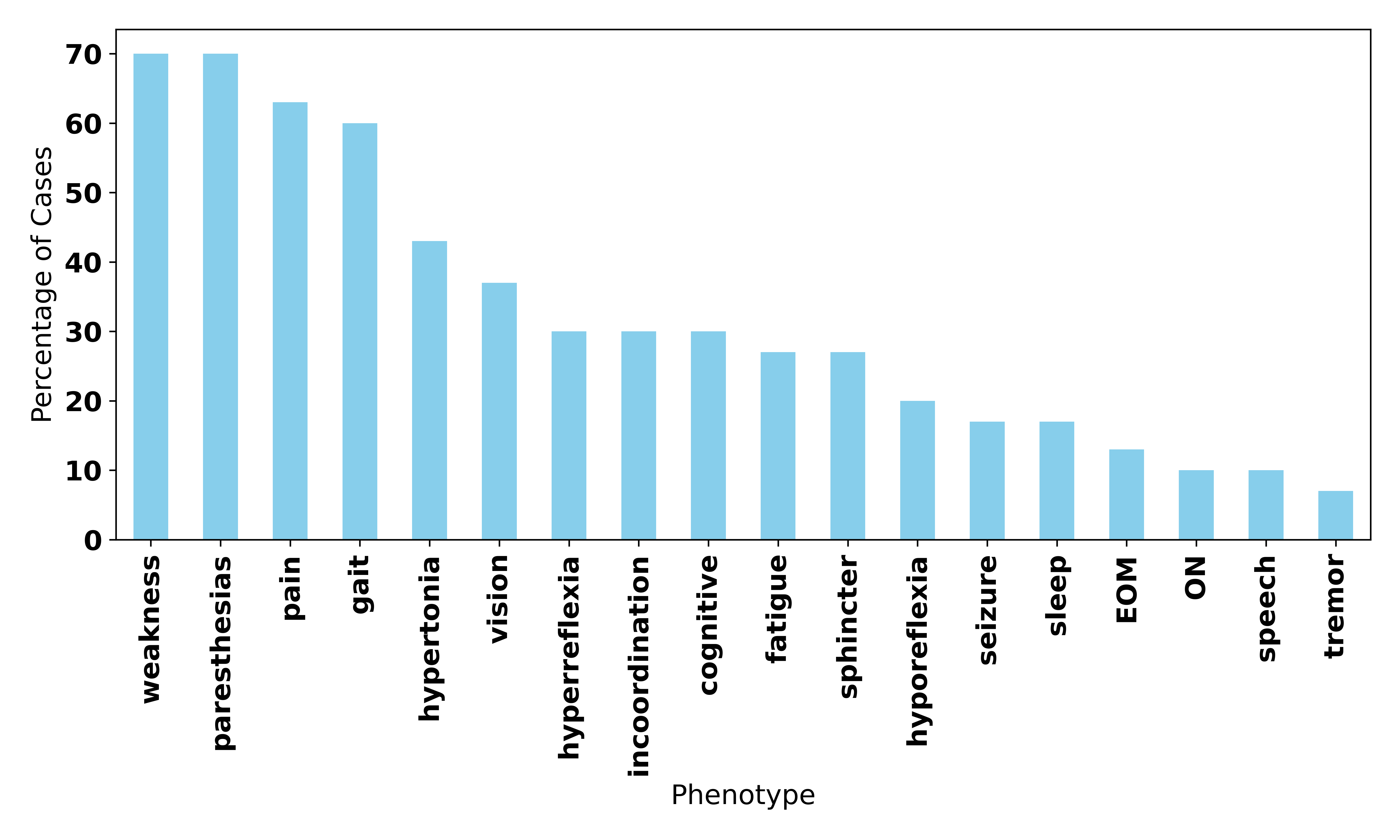}
     \captionsetup{font=footnotesize}
    \caption{
        \footnotesize Bar chart showing the frequency of each sign or symptom in the test set of patient notes. All patient notes had a diagnosis of multiple sclerosis (ICD-10 code G35). Each sign or symptom was coded binary (present or absent) regardless of the number of occurrences in the note. The most common signs and symptoms were weakness, paresthesias, pain, and gait. 
    }
    \label{fig:histogram}
\end{figure}
\begin{figure}
    \centering
    \includegraphics[width =0.49\textwidth]{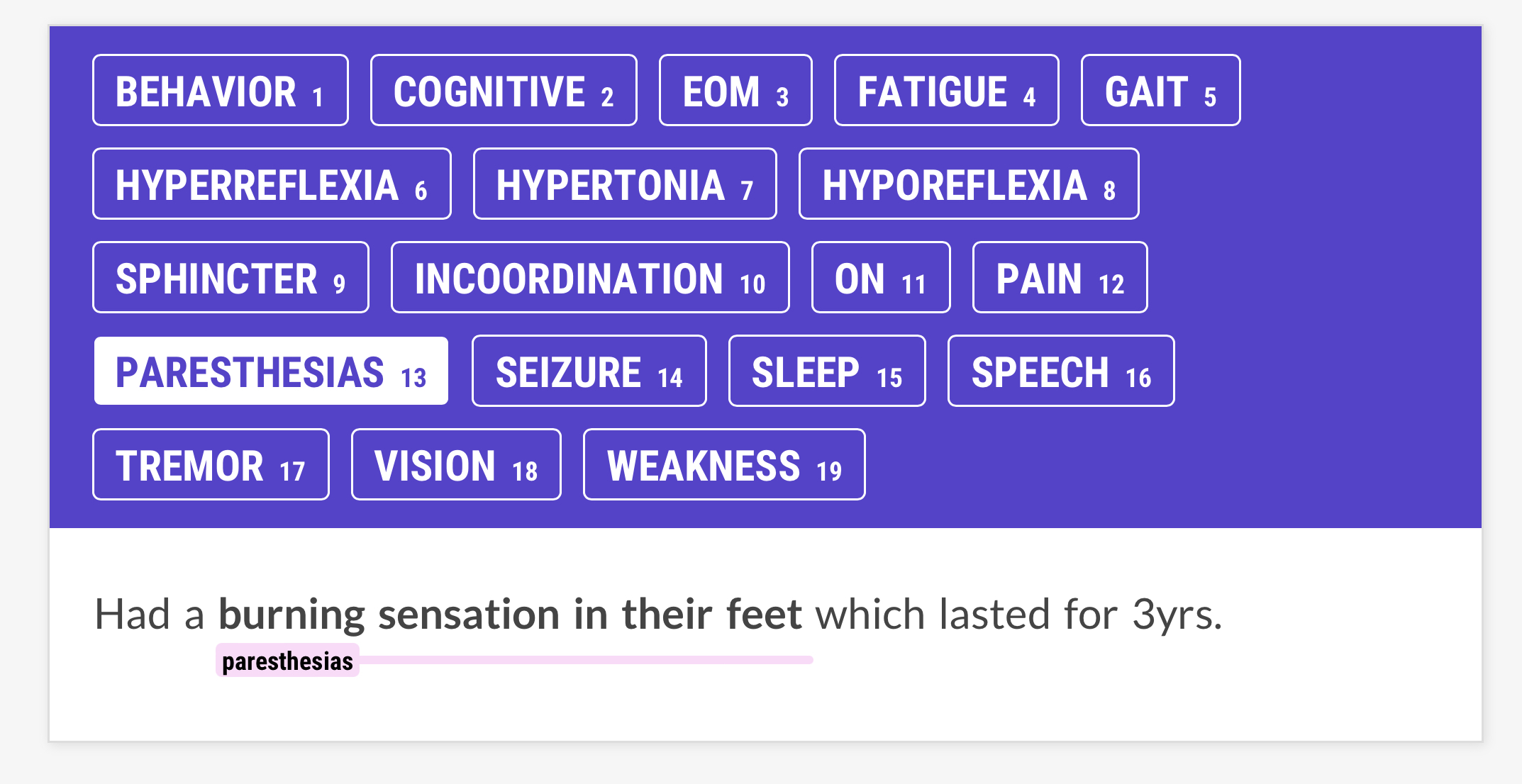}
\caption{Annotation screen for \textbf{Prodigy} using the manual.spancat recipe to label text spans (Explosion AI). The annotator has the choice of 19 neurological phenotype labels and has chosen \textbf{paresthesias} to label text span \textit{burning sensation in their feet}.}
    \label{fig:Prodigy}
\end{figure}

\section{Methods}
\textit{Data acquisition}. Physician narratives in the electronic medical records of neurology patients with a diagnosis of multiple sclerosis (ICD-10 code = G35) seen at the Neurology Clinic of the University of Illinois at Chicago between 2019 and 2022 were downloaded as a CSV file from the REDCap (Research Electronic Data Capture) system. The first physician progress note that was not a telehealth visit was analyzed for each research subject. Discharge summaries and admission notes were excluded from the analysis, and notes were deidentified. The Institutional Review Board of the University of Illinois at Chicago approved data collection and use. Five hundred forty-seven records, one for each unique research subject, were available for analysis. Thirty records were segregated to create the test set and excluded from training.

\textit{Phenotyping by human annotator}. The human annotator used the Prodigy annotation tool (manual.spancat recipe, Explosion AI, Berlin) to annotate physician notes with one of nineteen labels (Figure \ref{fig:Prodigy}). The human annotator was given the same instructions as  GPT-4 (Box 1).

\textit{Phenotyping by NimbleMiner}.  NimbleMiner \cite{nimbleminer} is a software tool for medical concept recognition in clinical texts that is implemented in R (the latest version is implemented in R version 4.0.2 \cite{R4.0.2}). Because NimbleMiner uses a machine learning classifier and word embeddings to find medical concepts, we referred to it as a hybrid NLP model. Word embeddings (word2vec) convert target terms into a lexicon of similar terms called \textit{simclins}.  With simclins as a target, NimbleMiner uses machine learning classifiers to search for matching phrases in clinical narratives. NimbleMiner follows a positive-only label learning strategy and excludes negated concepts.

\begin{figure*}
    \centering
    \includegraphics[width =0.80\textwidth]{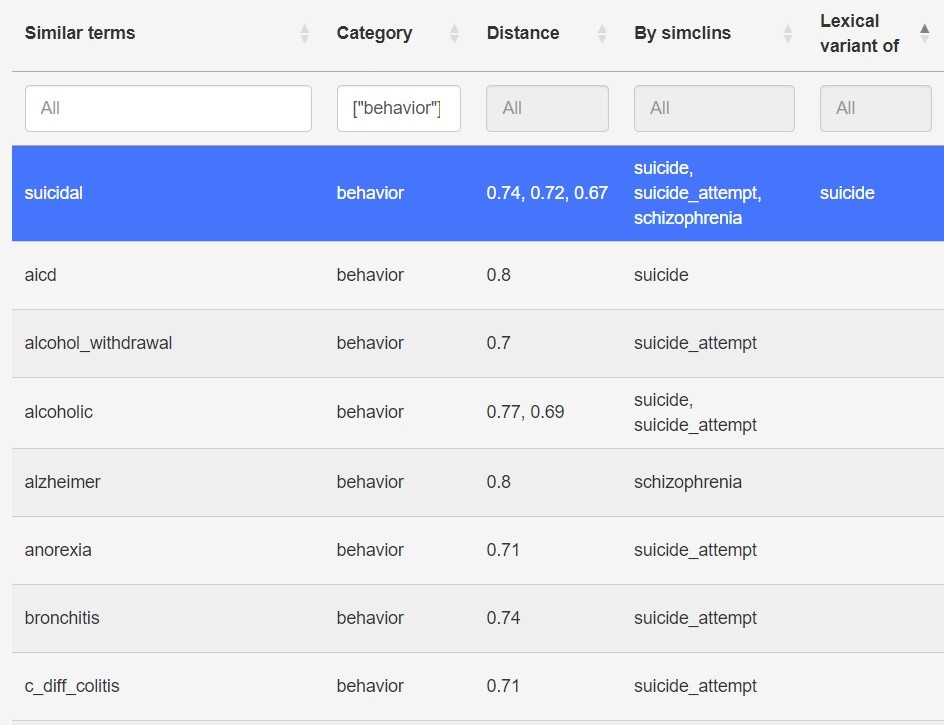}
\caption{Simclin Explorer screen for \textbf{NimbleMiner} for the annotation of the feature \textbf{behavior}. The blue highlighted row indicates a simclin selected for inclusion in the final simclins list. The gray rows represent terms that were marked irrelevant.}
    \label{fig:Simclin_explorer_screen}
\end{figure*}

An initial list of multiple sclerosis signs and symptoms of multiple sclerosis was supplemented by text spans found in neurology notes (for sample spans, see Box 2). The final list of seed terms was passed to NimbleMiner to create a lexicon of simclins using the native distance metric in Nimbleminer. Terms and phrases within a distance of x $>$ 0.6 from any seed term were individually evaluated as relevant or irrelevant. Relevant terms were retained on the simclins list (Fig. \ref{fig:Simclin_explorer_screen}).  Negation terms were classified as pre-negations and post-negations. Pre-negation terms were terms that appeared before a text span and negated the span (e.g., \textit{\textbf{no sign of} weakness}). Post-negations were terms that appeared after a particular text and negated the span (e.g., \textit{weakness \textbf{negative}}). NimbleMinder was trained to recognize 19 categories of neurological signs and symptoms of multiple sclerosis (Fig. \ref{fig:histogram} and \ref{fig:Prodigy}). The native support vector machine (SVM) classifier within Nimbleminer evaluated whether a simclin appeared in a note and outputted \say{true} or \say{false} on a note-by-note basis. Results were binarized as a \say{1} for true and \say{0} for false.

\textit{Phenotyping by  GPT-4}. The instructions for phenotyping the physician notes were passed to GPT-4 \cite{OpenAI2024ChatGPT} as a prompt (Box 1). As a design feature of  GPT-4, it retaina the instructions in \say{short-term memory} throughout a  session and does not need to have them re-stated when multiple cases are analyzed. The output from  GPT-4 was a format that allowed easy analysis (Box 3).

\begin{mdframed}[
  frametitle={Box 2: Example training phrases passed to NimbleMiner to generate simclins (similar clinical terms) by word embedding},
  frametitlebackgroundcolor=gray!30,
  backgroundcolor=gray!20,
  frametitlerule=true,
  frametitlealignment=\raggedright,
  innertopmargin=10pt,
  innerbottommargin=10pt,
  innerleftmargin=10pt,
  innerrightmargin=10pt,
  roundcorner=10pt,
  font=\footnotesize
]
\begin{tabular}{ll}
    \textbf{Span} & \textbf{Phenotype} \\
    Depressed mood & behavior \\
    difficulty thinking clearly & cognitive \\
    double vision & EOM \\
    worsening fatigue & fatigue \\
    gait instability & gait \\
    diffuse hyperreflexia & hyperreflexia \\
    spasticity worse & hypertonia \\
    pale temporal nerve OS & ON \\
    Numbness or tingling + & paresthesias \\
    Neurogenic bladder & sphincter \\
    blurry vision & vision \\
    endorses left-sided weakness & weakness \\
    electric shock & pain \\
\end{tabular}.
\end{mdframed}

\begin{table}[!ht]
    \centering
    \begin{threeparttable}
    \caption{Model Performance Metrics For NimbleMiner and  GPT-4}
    \begin{tabular}{|l|l|l|l|l|l|}
    \hline
        Implementation & Accuracy & Precision & Recall & Specificity&F1 \\ \hline
        NimbleMiner$^{1}$ & 0.87 & 0.82 & 0.81 & 0.88 & 0.78 \\ \hline
         GPT-4$^{2}$ & 0.85 & 0.79&0.72 & 0.91 & 0.73 \\ \hline
    \end{tabular}
    \label{Tab:Metrics}
    \begin{tablenotes}
      \small
      \item[1] NimbleMiner with word embedding and (support vector machine) SVM classifier \cite{nimbleminer}.
      \item[2] GPT-4 is a large language model (LLM) \cite{OpenAI2024ChatGPT}.
    \end{tablenotes}
    \end{threeparttable}
\end{table}

\textit{Calculation of Performance Metrics}.  Results of the phenotyping by the human annotator, NimbleMiner, and GPT-4 were converted to a 30 x 20 array where each column was a phenotype (Fig. \ref{fig:Prodigy}) and each row was a physician note. Cells were coded \say{0} or \say{1} depending on whether the phenotype was absent or present. Accuracy, precision, recall, and F1 were calculated by standard methods \cite{velupillai2009developing, PrecisionAndRecallWiki}. Since the classification problem had nineteen labels, performance metrics were averaged across all labels.

\section{Results}
\label{s:results}
 Ground-truth labels for the thirty neurological test case notes were created using the spans.manual recipe from Prodigy (Explosion AI, Berlin). The human annotator was given the same instructions as GPT-4 (Box 1) and had access to 19 phenotype labels (Fig. \ref{fig:Prodigy}).  The test notes produced 2892 lines of text, and each was displayed separately on an annotation screen (Fig. \ref{fig:Prodigy}).  The annotator found 606 ground truth labels. Labels were distributed as shown in Fig. \ref{fig:histogram}. The most common phenotypes were weakness, paresthesias, pain, and gait abnormalities. Performance metrics for finding the ground truth labels in the physician notes were calculated for the NimbleMiner and  GPT-4 models (Table \ref{Tab:Metrics}).  NimbleMiner and GPT-4 demonstrated high accuracy, precision, specificity, and recall scores.  
\begin{mdframed}[
  frametitle={Box 3: Phenotyping of Case 170 by GPT-4},
  frametitlebackgroundcolor=gray!30, 
  backgroundcolor=gray!20, 
  frametitlerule=true,
  frametitlealignment=\centering,
  innertopmargin=10pt,
  innerbottommargin=10pt,
  innerleftmargin=10pt,
  innerrightmargin=10pt,
  roundcorner=10pt,
  font=\footnotesize 
]
Behavior: None\\
Cognitive: Memory loss (feels forgetful, needs to write things down)\\
EOM: None\\
Fatigue: None\\
Gait: Difficulty walking, unable to walk in a straight line\\
Hyperreflexia: None\\
Hypertonia: None\\
Hyporeflexia: None\\
Sphincter: Urinary urgency\\
Incoordination: None\\
ON: None\\
Pain: None\\
Paresthesias: None\\
Seizure: Questionable history of seizure\\
Sleep: None\\
Speech: None\\
Tremor: None\\
Vision: None\\
Weakness: Chronic right-sided weakness
\end{mdframed}
\section{Discussion}
\label{s:discussion}
Identifying the signs and symptoms of patients in clinical notes is a crucial component of precision medicine. To support big data and precision medicine, these signs and symptoms are mapped to concepts in standardized terminologies and converted to machine-readable codes. This process has been named \textit{deep phenotyping} \cite{robinson2012deep}. Although deep phenotyping can be done manually, the large number of physician notes in electronic records requires automated methods \cite{hier_metamap}. We have estimated that a medium-sized hospital with 300 beds can generate as many as 1.5 million physician notes annually \cite{oommen2023inter} 

The high throughput phenotyping of physician notes has proven difficult by natural language processing methods for several reasons:
\begin{enumerate}
    \item Many medical concepts have multiple synonyms (synonymy). For example, dystaxia, ataxia, and dysmetria all describe the neurological concept of uncoordinated movements.
    \item Word meanings in medicine may vary depending on context (polysemy). For example, a neurologist may use a pin to test sensation, and an orthopedic surgeon may use a pin to repair a fracture.
    \item Physicians may use colloquialisms and figures of speech such as \say{upgoing toe} in place of Babinski sign or  \say{patient is feeling blue} instead of saying that the patient is depressed.
    \item Physician notes are marred by irregular abbreviations, misspellings, and typographical errors.
    \item Physicians may substitute a concept description for a concept name such as substituting \say{patient has slurring of their speech} for dysarthria.
    \item Physician notes enumerate many negative and normal findings which may be mistaken for positive findings such as confusing \say{no weakness} with weakness.
\end{enumerate}

Nonetheless, progress has been made. The hybrid model that combined an SVM classifier with word embeddings (NimbleMiner) and the large language model (GPT-4) performed well on the neurology note phenotyping task with accuracies of 0.87 and 0.85, respectively (Table \ref{Tab:Metrics}). These high levels of accuracy are impressive given that the level of agreement between human annotators reaches a ceiling at a $\kappa$ $\approx$ $0.90$ \cite{oommen2023inter}. 

NimbleMiner has several advantages for high throughput phenotyping. It is fast, relatively easy to implement, transparent, and performs with high recall, precision, and specificity when properly configured with appropriate simclins. It uses a positive-only labeling strategy compatible with phenotyping, where negative and normal findings can be ignored. On the other hand, if the simclins (similar clinical terms) are not carefully formulated, recall (sensitivity) may suffer. An iterative approach is often needed with additional simclins added to the lexicon until an acceptable level of recall is reached. A major drawback of NimbleMiner is that the simclins and lexicon must be reconfigured for each new phenotyping application (see Fig. \ref{fig:Simclin_explorer_screen}).  A validation dataset with ground truth annotations is needed to ensure the model functions as desired.

The advantages of using large language models like GPT-4 for high throughput phenotyping are obvious. The model is easy to configure (Box 1).  Furthermore, no training set is needed. Accuracy appears as good as or better than other methods. Output is intelligible and easy to configure (Box 3). Concept extraction within the free medical text is a two-step process that involves identification (finding the relevant concepts) and mapping (assigning each concept to a class in a defined terminology) \cite {krauthammer2004term}. GPT-4 can identify neurological concepts in free text and map them to one of the nineteen phenotype categories without additional training (Boxes 1 and 2).  Nonetheless, Large language models like GPT-4 are computationally expensive.

Although our results with GPT-4 and NimbleMiner are encouraging, confirmation of these results with a larger and more diverse corpus of physician notes is needed. Given the power and simplicity of performing high throughput phenotyping with large language models, these models will likely become the dominant method \cite{omiye2023large, thompson2023large}.
\nocite{}
\bibliographystyle{IEEEtran}
\bibliography{references}

\begin{thebibliography}{10}
\providecommand{\url}[1]{#1}
\csname url@samestyle\endcsname
\providecommand{\newblock}{\relax}
\providecommand{\bibinfo}[2]{#2}
\providecommand{\BIBentrySTDinterwordspacing}{\spaceskip=0pt\relax}
\providecommand{\BIBentryALTinterwordstretchfactor}{4}
\providecommand{\BIBentryALTinterwordspacing}{\spaceskip=\fontdimen2\font plus
\BIBentryALTinterwordstretchfactor\fontdimen3\font minus \fontdimen4\font\relax}
\providecommand{\BIBforeignlanguage}[2]{{%
\expandafter\ifx\csname l@#1\endcsname\relax
\typeout{** WARNING: IEEEtran.bst: No hyphenation pattern has been}%
\typeout{** loaded for the language `#1'. Using the pattern for}%
\typeout{** the default language instead.}%
\else
\language=\csname l@#1\endcsname
\fi
#2}}
\providecommand{\BIBdecl}{\relax}
\BIBdecl

\bibitem{sahu2022artificial}
M.~Sahu, R.~Gupta, R.~K. Ambasta, and P.~Kumar, ``Artificial intelligence and machine learning in precision medicine: A paradigm shift in big data analysis,'' \emph{Progress in Molecular Biology and Translational Science}, vol. 190, no.~1, pp. 57--100, 2022.

\bibitem{afzal2020precision}
M.~Afzal, S.~R. Islam, M.~Hussain, and S.~Lee, ``Precision medicine informatics: principles, prospects, and challenges,'' \emph{IEEE Access}, vol.~8, pp. 13\,593--13\,612, 2020.

\bibitem{robinson2012deep}
P.~N. Robinson, ``Deep phenotyping for precision medicine,'' \emph{Human mutation}, vol.~33, no.~5, pp. 777--780, 2012.

\bibitem{hier2022focused}
D.~Hier, R.~Yelugam, S.~Azizi, and D.~Wunsch~III, ``A focused review of deep phenotyping with examples from neurology,'' \emph{Eur Sci J}, vol.~18, pp. 4--19, 2022.

\bibitem{hier2022dc}
D.~Hier, R.~Yelugam, S.~Azizi, M.~Carrithers, and I.~Wunsch, ``Dc. high throughput neurological phenotyping with metamap,'' \emph{Eur Sci J}, vol.~18, pp. 37--49, 2022.

\bibitem{mir2019high}
R.~R. Mir, M.~Reynolds, F.~Pinto, M.~A. Khan, and M.~A. Bhat, ``High-throughput phenotyping for crop improvement in the genomics era,'' \emph{Plant Science}, vol. 282, pp. 60--72, 2019.

\bibitem{gehan2017high}
M.~A. Gehan and E.~A. Kellogg, ``High-throughput phenotyping,'' \emph{American journal of botany}, vol. 104, no.~4, pp. 505--508, 2017.

\bibitem{alzoubi2019review}
H.~Alzoubi, R.~Alzubi, N.~Ramzan, D.~West, T.~Al-Hadhrami, and M.~Alazab, ``A review of automatic phenotyping approaches using electronic health records,'' \emph{Electronics}, vol.~8, no.~11, p. 1235, 2019.

\bibitem{shivade2014review}
C.~Shivade, P.~Raghavan, E.~Fosler-Lussier, P.~J. Embi, N.~Elhadad, S.~B. Johnson, and A.~M. Lai, ``A review of approaches to identifying patient phenotype cohorts using electronic health records,'' \emph{Journal of the American Medical Informatics Association}, vol.~21, no.~2, pp. 221--230, 2014.

\bibitem{pathak2013electronic}
J.~Pathak, A.~N. Kho, and J.~C. Denny, ``Electronic health records-driven phenotyping: challenges, recent advances, and perspectives,'' \emph{Journal of the American Medical Informatics Association}, vol.~20, no.~e2, pp. e206--e211, 2013.

\bibitem{krauthammer2004term}
M.~Krauthammer and G.~Nenadic, ``Term identification in the biomedical literature,'' \emph{Journal of biomedical informatics}, vol.~37, no.~6, pp. 512--526, 2004.

\bibitem{Bird}
S.~Bird, E.~Klein, and E.~Loper, \emph{Natural Language Processing with Python}.\hskip 1em plus 0.5em minus 0.4em\relax O'Reilly Media, Sebastopol, CA, 2009.

\bibitem{luque2019advanced}
C.~Luque, J.~M. Luna, M.~Luque, and S.~Ventura, ``An advanced review on text mining in medicine,'' \emph{Wiley Interdisciplinary Reviews: Data Mining and Knowledge Discovery}, vol.~9, no.~3, p. e1302, 2019.

\bibitem{ahmad2023review}
P.~N. Ahmad, A.~M. Shah, and K.~Lee, ``A review on electronic health record text-mining for biomedical name entity recognition in healthcare domain,'' in \emph{Healthcare}, vol.~11, no.~9.\hskip 1em plus 0.5em minus 0.4em\relax MDPI, 2023, p. 1268.

\bibitem{eltyeb2014chemical}
S.~Eltyeb and N.~Salim, ``Chemical named entities recognition: a review on approaches and applications,'' \emph{Journal of cheminformatics}, vol.~6, no.~1, pp. 1--12, 2014.

\bibitem{quimbaya2016named}
A.~P. Quimbaya, A.~S. M{\'u}nera, R.~A.~G. Rivera, J.~C.~D. Rodr{\'\i}guez, O.~M.~M. Velandia, A.~A.~G. Pe{\~n}a, and C.~Labb{\'e}, ``Named entity recognition over electronic health records through a combined dictionary-based approach,'' \emph{Procedia Computer Science}, vol. 100, pp. 55--61, 2016.

\bibitem{hirschman2002rutabaga}
L.~Hirschman, A.~A. Morgan, and A.~S. Yeh, ``Rutabaga by any other name: extracting biological names,'' \emph{Journal of Biomedical Informatics}, vol.~35, no.~4, pp. 247--259, 2002.

\bibitem{uzuner20112010}
{\"O}.~Uzuner, B.~R. South, S.~Shen, and S.~L. DuVall, ``2010 i2b2/va challenge on concepts, assertions, and relations in clinical text,'' \emph{Journal of the American Medical Informatics Association}, vol.~18, no.~5, pp. 552--556, 2011.

\bibitem{aronson2010overview}
A.~R. Aronson and F.-M. Lang, ``An overview of metamap: historical perspective and recent advances,'' \emph{Journal of the American Medical Informatics Association}, vol.~17, no.~3, pp. 229--236, 2010.

\bibitem{savova2010mayo}
G.~K. Savova, J.~J. Masanz, P.~V. Ogren, J.~Zheng, S.~Sohn, K.~C. Kipper-Schuler, and C.~G. Chute, ``Mayo clinical text analysis and knowledge extraction system ({cTAKES}): architecture, component evaluation and applications,'' \emph{Journal of the American Medical Informatics Association}, vol.~17, no.~5, pp. 507--513, 2010.

\bibitem{lample2016neural}
G.~Lample, M.~Ballesteros, S.~Subramanian, K.~Kawakami, and C.~Dyer, ``Neural architectures for named entity recognition,'' \emph{arXiv preprint arXiv:1603.01360}, 2016.

\bibitem{chiu2016named}
J.~P. Chiu and E.~Nichols, ``Named entity recognition with bidirectional {LSTM-CNNs},'' \emph{Transactions of the Association for Computational Linguistics}, vol.~4, pp. 357--370, 2016.

\bibitem{habibi2017deep}
M.~Habibi, L.~Weber, M.~Neves, D.~L. Wiegandt, and U.~Leser, ``Deep learning with word embeddings improves biomedical named entity recognition,'' \emph{Bioinformatics}, vol.~33, no.~14, pp. i37--i48, 2017.

\bibitem{gehrmann2018comparing}
S.~Gehrmann, F.~Dernoncourt, Y.~Li, E.~T. Carlson, J.~T. Wu, J.~Welt, J.~Foote~Jr, E.~T. Moseley, D.~W. Grant, P.~D. Tyler \emph{et~al.}, ``Comparing deep learning and concept extraction based methods for patient phenotyping from clinical narratives,'' \emph{PloS one}, vol.~13, no.~2, p. e0192360, 2018.

\bibitem{arbabi2019identifying}
A.~Arbabi, D.~R. Adams, S.~Fidler, M.~Brudno \emph{et~al.}, ``Identifying clinical terms in medical text using ontology-guided machine learning,'' \emph{JMIR medical informatics}, vol.~7, no.~2, p. e12596, 2019.

\bibitem{devlin2018bert}
J.~Devlin, M.-W. Chang, K.~Lee, and K.~Toutanova, ``{BERT}: Pre-training of deep bidirectional transformers for language understanding,'' \emph{arXiv preprint arXiv:1810.04805}, 2018.

\bibitem{vaswani2017attention}
A.~Vaswani, N.~Shazeer, N.~Parmar, J.~Uszkoreit, L.~Jones, A.~N. Gomez, {\L}.~Kaiser, and I.~Polosukhin, ``Attention is all you need,'' in \emph{Advances in neural information processing systems}, 2017, pp. 5998--6008.

\bibitem{zhu2021utilizing}
R.~Zhu, X.~Tu, and J.~X. Huang, ``Utilizing {BERT} for biomedical and clinical text mining,'' in \emph{Data Analytics in Biomedical Engineering and Healthcare}.\hskip 1em plus 0.5em minus 0.4em\relax Elsevier, 2021, pp. 73--103.

\bibitem{yu2019biobert}
X.~Yu, W.~Hu, S.~Lu, X.~Sun, and Z.~Yuan, ``Biobert based named entity recognition in electronic medical record,'' \emph{2019 10th international conference on information technology in medicine and education (ITME)}, pp. 49--52, 2019.

\bibitem{lee2020biobert}
J.~Lee, W.~Yoon, S.~Kim, D.~Kim, S.~Kim, C.~H. So, and J.~Kang, ``Biobert: a pre-trained biomedical language representation model for biomedical text mining,'' \emph{Bioinformatics}, vol.~36, no.~4, pp. 1234--1240, 2020.

\bibitem{ji2020bert}
Z.~Ji, Q.~Wei, and H.~Xu, ``Bert-based ranking for biomedical entity normalization,'' \emph{AMIA Summits on Translational Science Proceedings}, vol. 2020, p. 269, 2020.

\bibitem{yan2023large}
C.~Yan, H.~Ong, M.~Grabowska, M.~Krantz, W.-C. Su, A.~Dickson, J.~F. Peterson, Q.~Feng, D.~M. Roden, C.~M. Stein \emph{et~al.}, ``Large language models facilitate the generation of electronic health record phenotyping algorithms,'' \emph{medRxiv}, pp. 2023--12, 2023.

\bibitem{yang2023enhancing}
J.~Yang, C.~Liu, W.~Deng, D.~Wu, C.~Weng, Y.~Zhou, and K.~Wang, ``Enhancing phenotype recognition in clinical notes using large language models: Phenobcbert and phenogpt,'' \emph{Patterns}, 2023.

\bibitem{wang2023fine}
A.~Wang, C.~Liu, J.~Yang, and C.~Weng, ``Fine-tuning large language models for rare disease concept normalization,'' \emph{bioRxiv}, pp. 2023--12, 2023.

\bibitem{nimbleminer}
M.~Topaz, ``Nimbleminer: A novel multi-lingual text mining application,'' \emph{MEDINFO 2019: Health and Wellbeing e-Networks for All}, pp. 1608--1609, 2019.

\bibitem{R4.0.2}
\BIBentryALTinterwordspacing
{R Core Team}, \emph{R: A Language and Environment for Statistical Computing}, R Foundation for Statistical Computing, Vienna, Austria, 2020. [Online]. Available: \url{https://www.R-project.org/}
\BIBentrySTDinterwordspacing

\bibitem{OpenAI2024ChatGPT}
\BIBentryALTinterwordspacing
OpenAI, ``Chatgpt (4),'' Large language model, 2024. [Online]. Available: \url{https://chat.openai.com}
\BIBentrySTDinterwordspacing

\bibitem{velupillai2009developing}
S.~Velupillai, H.~Dalianis, M.~Hassel, and G.~H. Nilsson, ``Developing a standard for de-identifying electronic patient records written in swedish: precision, recall and f-measure in a manual and computerized annotation trial,'' \emph{International journal of medical informatics}, vol.~78, no.~12, pp. e19--e26, 2009.

\bibitem{PrecisionAndRecallWiki}
\BIBentryALTinterwordspacing
Wikipedia, ``Precision and recall,'' accessed: [January 29, 2024]. [Online]. Available: \url{https://en.wikipedia.org/wiki/Precision_and_recall}
\BIBentrySTDinterwordspacing

\bibitem{hier_metamap}
\BIBentryALTinterwordspacing
D.~B. Hier, R.~Yelugam, S.~Azizi, M.~D. Carrithers, and D.~C. Wunsch~II, ``High throughput neurological phenotyping with {MetaMap},'' \emph{European Scientific Journal}, vol.~18, pp. 37--49, 2022, accessed August 12, 2022. [Online]. Available: \url{https://doi.org/10.190444/esj.2022.v18n4p37}
\BIBentrySTDinterwordspacing

\bibitem{oommen2023inter}
C.~Oommen, Q.~Howlett-Prieto, M.~D. Carrithers, and D.~B. Hier, ``Inter-rater agreement for the annotation of neurologic signs and symptoms in electronic health records,'' \emph{Frontiers in Digital Health}, vol.~5, p. 1075771, 2023.

\bibitem{omiye2023large}
J.~A. Omiye, H.~Gui, S.~J. Rezaei, J.~Zou, and R.~Daneshjou, ``Large language models in medicine: the potentials and pitfalls,'' \emph{arXiv preprint arXiv:2309.00087}, 2023.

\bibitem{thompson2023large}
W.~E. Thompson, D.~M. Vidmar, J.~K. De~Freitas, J.~M. Pfeifer, B.~K. Fornwalt, R.~Chen, G.~Altay, K.~Manghnani, A.~C. Nelsen, K.~Morland \emph{et~al.}, ``Large language models with retrieval-augmented generation for zero-shot disease phenotyping,'' \emph{arXiv preprint arXiv:2312.06457}, 2023.

\end{thebibliography}
\end{document}